\documentclass{article}

\usepackage[preprint]{neurips_2025}

\usepackage[utf8]{inputenc} 
\usepackage[T1]{fontenc}    
\usepackage{hyperref}       
\usepackage{url}            
\usepackage{booktabs}       
\usepackage{amsfonts}       
\usepackage{nicefrac}       
\usepackage{microtype}      
\usepackage{xcolor}         
\usepackage{graphicx}
\usepackage{amsmath}

\title{Cut2Next: \\Generating Next Shot via In-Context Tuning}

\author{%
  Jingwen He\textsuperscript{1,2}  Hongbo Liu\textsuperscript{2}  Jiajun Li\textsuperscript{3}  \textbf{Ziqi Huang}\textsuperscript{3}  \textbf{Yu Qiao}\textsuperscript{2}  \textbf{Wanli Ouyang}\textsuperscript{1}\textsuperscript{$\dagger$}  \textbf{Ziwei Liu}\textsuperscript{3}\textsuperscript{$\dagger$} \\
  \textsuperscript{1}The Chinese University of Hong Kong,  \textsuperscript{2}Shanghai Artificial Intelligence Laboratory, \\ \textsuperscript{3}S-Lab, Nanyang Technological University \\ \textsuperscript{$\dagger$}Corresponding authors.\\
  Project page: \href{https://vchitect.github.io/Cut2Next-project/}{\texttt{https://vchitect.github.io/Cut2Next-project/}}
}

\begin{document}

\maketitle

\begin{figure}[h!] 
  \centering
  \includegraphics[width=\textwidth]{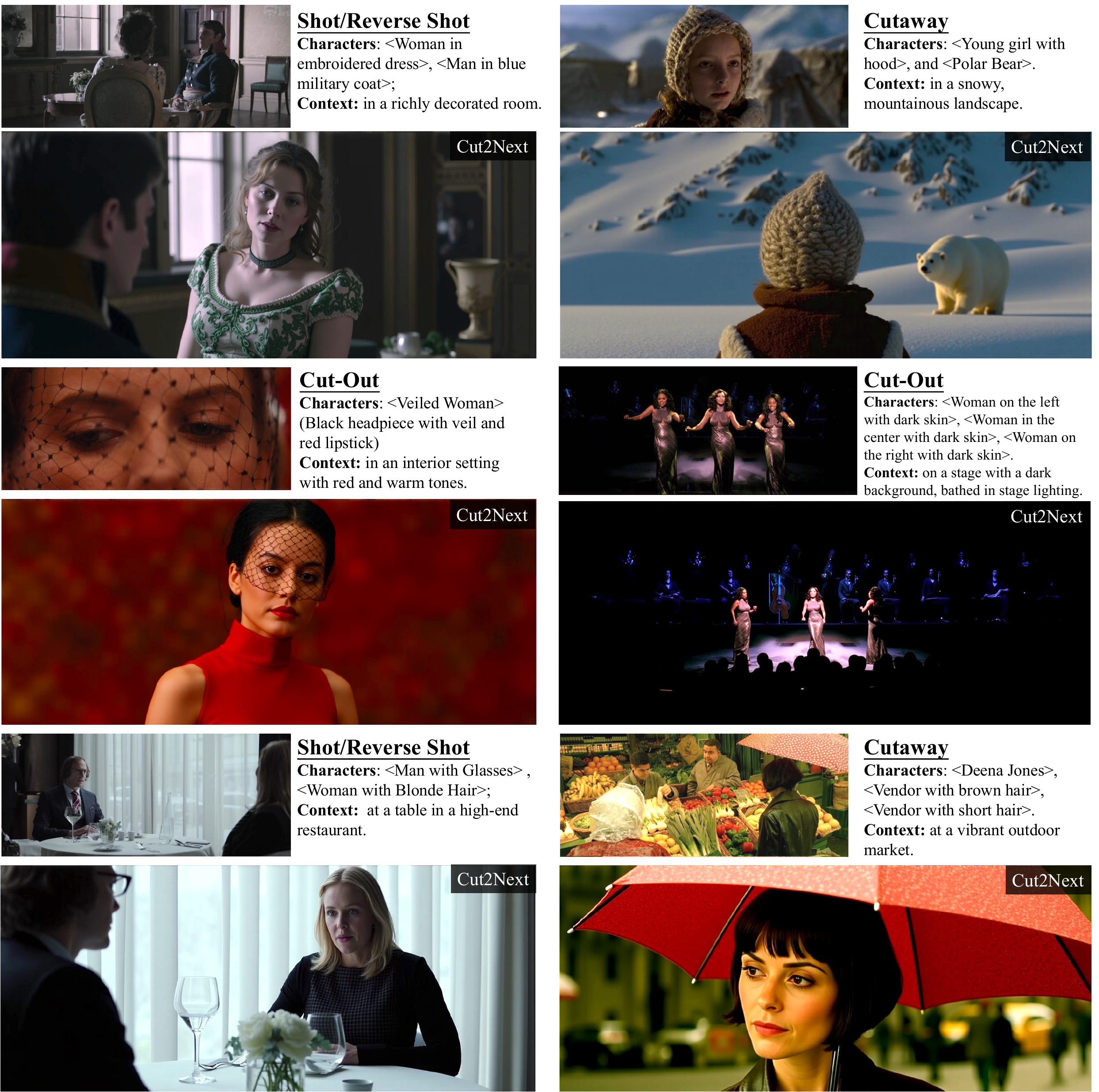} 
  \caption{\textbf{Cut2Next demonstrating versatile Next Shot Generation}. The model produces cinematically coherent subsequent shots (bottom) adhering to diverse editing patterns (e.g., Shot/Reverse Shot, Cut-Out, Cutaway) specified alongside the input shots (upper).}
  \label{fig:teaser} 
\end{figure}

\begin{abstract}
 Effective multi-shot generation demands purposeful, film-like transitions and strict cinematic continuity. Current methods, however, often prioritize basic visual consistency, neglecting crucial editing patterns (e.g., shot/reverse shot, cutaways) that drive narrative flow for compelling storytelling. This yields outputs that may be visually coherent but lack narrative sophistication and true cinematic integrity.
To bridge this, we introduce \textbf{Next Shot Generation (NSG)}: synthesizing a subsequent, high-quality shot that critically conforms to professional editing patterns while upholding rigorous cinematic continuity. Our framework, \textbf{Cut2Next}, leverages a Diffusion Transformer (DiT). It employs in-context tuning guided by a novel \textit{Hierarchical Multi-Prompting} strategy. This strategy uses \textit{Relational Prompts} to define overall context and inter-shot editing styles. \textit{Individual Prompts} then specify per-shot content and cinematographic attributes. Together, these guide Cut2Next to generate cinematically appropriate next shots. Architectural innovations, \textit{Context-Aware Condition Injection (CACI)} and \textit{Hierarchical Attention Mask (HAM)}, further integrate these diverse signals without introducing new parameters.
We construct RawCuts (large-scale) and CuratedCuts (refined) datasets, both with hierarchical prompts, and introduce CutBench for evaluation. Experiments show Cut2Next excels in visual consistency and text fidelity. Crucially, user studies reveal a strong preference for Cut2Next, particularly for its adherence to intended editing patterns and overall cinematic continuity, validating its ability to generate high-quality, narratively expressive, and cinematically coherent subsequent shots.
\end{abstract}

\section{Introduction}

Video generation has advanced remarkably. Models like Sora \cite{sora} and Kling \cite{kling} synthesize photorealistic single-shot videos, built upon large-scale data and scalable architectures like Diffusion Transformers (DiT) \cite{dit}. Following these successes, the academic community is increasingly focusing on narrative video generation \cite{lct,mask2dit,shotadapter,xiao2025videoauteur}. This aims to create videos composed of multiple interconnected shots, akin to films. 

Several approaches address multi-shot video generation. One prominent strategy is storyboard generation \cite{storydiffusion, vgot, one_prompt_one_story,tewel2024training}: text-to-image models first create keyframes, which image-to-video models then animate to construct the full video. Another approach is to directly tackle multi-shot video generation by training models on extensive video datasets \cite{lct, mask2dit, shotadapter}. Both these research lines primarily aim for diverse content in long-form videos with rich plotlines. In contrast, a distinct research line prioritizes individual shot quality and basic consistency within sequences. For instance, IC-LoRA \cite{iclora} leverages Flux~\cite{flux}'s in-context generation ability to generate high-resolution storyboards with consistent environments and characters. CineVerse \cite{cineverse} enhanced IC-LoRA with detailed annotations for user-controlled shot scale. SynCamMaster \cite{SynCamMaster} aims to synthesize 3D-consistent multi-angle, multi-scale shots, which are then edited together to enrich visual expression. While significantly advancing shot quality and basic consistency, these approaches often overlook the explicit modeling and enforcement of complex editing patterns crucial to professional narrative filmmaking.

\begin{figure*}[htbp]
\centering
\includegraphics[width=1.\linewidth]
{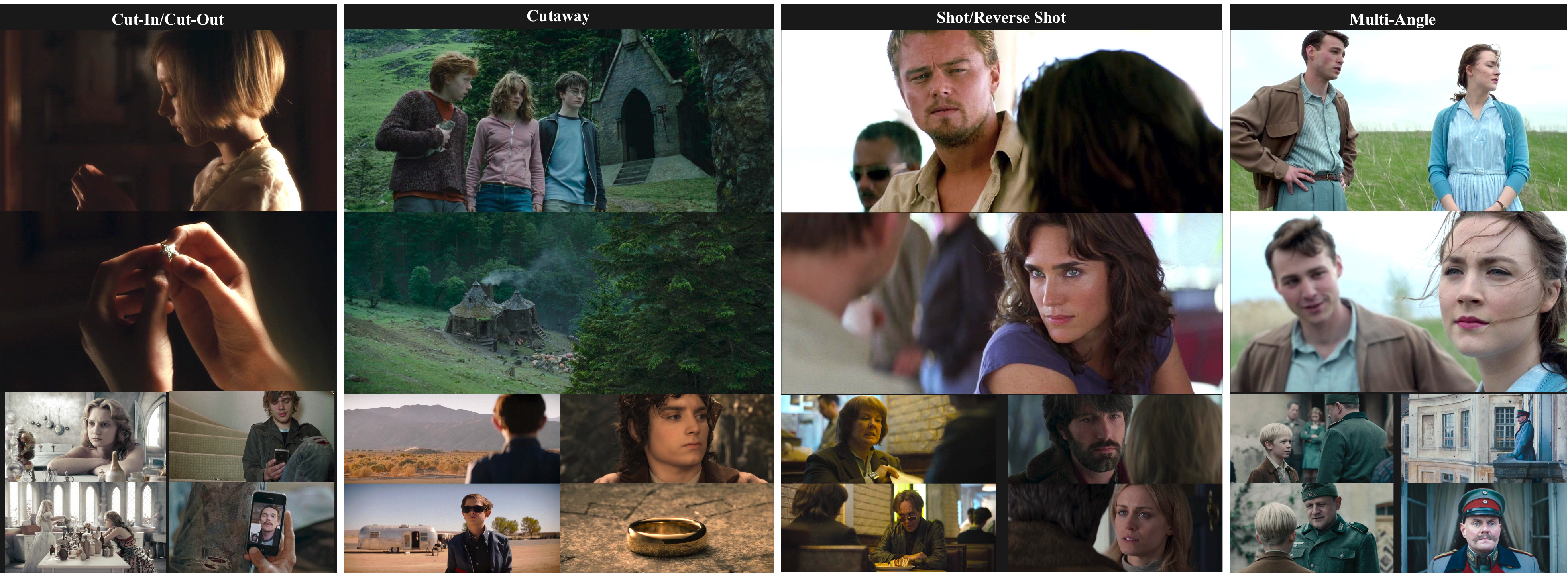}
  \caption{\textbf{Canonical cut-driven shot sequences} (from CuratedCuts), their narrative functions, and the generation difficulties.
\textbf{Cut-In/Cut-Out}: Emphasizes details or shifts focus; challenges models with drastic scale changes while maintaining subject consistency.
\textbf{Cutaway}: Provides external or subjective context; demands generating novel yet semantically related content.
\textbf{Shot/Reverse Shot}: Facilitates dialogue and reveals reactions; requires consistent character appearance and spatial logic across alternating viewpoints.
\textbf{Multi-Angle}: Offers varied viewpoints; requires consistent rendering across significant visual transformations.}
  \label{fig:some_cuts}
\end{figure*}

Our work addresses this crucial gap. We align with the pursuit of high-fidelity individual shots but place an emphasis on the fundamental cinematic concept of the "cut." In professionally edited movies, cuts are not arbitrary; they serve a distinct narrative purpose \cite{moviecuts}. Building on this, as illustrated in Figure~\ref{fig:some_cuts}, we focus on canonical cut-driven sequences: Shot/Reverse Shot for dialogue, Cut-In/Cut-Out for emphasis, Cutaways for context, and Multi-Angle for perspective shifts. Such sequences are vital for enhancing narrative capability and inter-shot consistency. Traditionally, filming these sequences is difficult and expensive. It requires shooting different angles one by one, repeated actor performances, and careful resetting of the scene and lights.

Based on these considerations, we define a new task: \textbf{Next Shot Generation (NSG)}. Given an existing shot, the goal is to generate a subsequent, highly coherent shot. This new shot must maintain character and environmental consistency while also adhering to established cinematic continuity principles and specific editing patterns.
This task of Next Shot Generation presents significant challenges:
First, it demands comprehensive cinematic continuity from complex inputs. Unlike typical subject-driven~\cite{ominicontrol,videobooth,cao2023masactrl,zhang2024flashface,gal2024lcm} tasks focused on isolated subject preservation, NSG requires consistency across numerous facets: character identity and appearance, spatial relationships, environmental details, lighting, color palettes, overall tone, and implied temporal progression. Achieving such multifaceted continuity from rich visual inputs is a foundational difficulty.
Second, NSG aims to replicate diverse editing patterns, introducing significant visual diversity. Examples include Shot/Reverse Shot (shifting to a character's converse view in dialogue), Cut-In (focusing on a detail for emphasis)/Cut-Out (returning to a wider context), Cutaway (providing external or subjective context), and Multi-Angle (cutting between different views of the same subject). The model must accurately synthesize these diverse visual outputs reflecting the specified editing pattern, while simultaneously upholding the comprehensive cinematic continuity.
Next Shot Generation uniquely balances this required shot diversity with strict, multi-faceted continuity, making it more advanced than simpler consistency tasks.

To achieve the demanding balance of Next Shot Generation (NSG), we first construct a comprehensive data foundation. This includes RawCuts, a large-scale dataset of adjacent shot pairs for foundational learning of visual transitions. Complementing this, CuratedCuts, a smaller, meticulously human-curated set, exemplifies the strong cinematic continuity and professional editing techniques crucial for NSG, serving fine-grained refinement. To guide our model in navigating the dual demands of continuity and diversity, we employ a Hierarchical Prompt Annotation scheme. This provides Relational Prompt to dictate inter-shot relationships and editing patterns, and Individual Prompts to detail per-shot content and cinematographic attributes.
Methodologically, we propose Cut2Next, built upon \texttt{FLUX.1-dev}~\cite{flux} using in-context tuning. To specifically address NSG's challenges, Cut2Next incorporates: (1) Our \textit{Context-Aware Condition Injection (CACI)} mechanism for nuanced integration of diverse conditional inputs. (2) A \textit{Hierarchical Attention Mask (HAM)} to orchestrate the intricate flow between visual and textual tokens. This approach allows Cut2Next to effectively utilize our rich annotations and generate high-quality, cinematically coherent next shots that masterfully balance diversity with continuity, all without introducing additional parameters to the base model.

To rigorously evaluate Cut2Next's capabilities in balancing these NSG demands, we introduce CutBench. This new benchmark features diverse movie shot images, each paired with our hierarchical prompts designed to elicit specific cinematic continuations that test both continuity and varied editing.
Our quantitative results demonstrate Cut2Next significantly outperforms baselines in generating visually coherent and textually aligned subsequent shots. Crucially, to evaluate the critical balance—cinematic continuity and intended editing patterns—we conducted extensive user studies. These studies not only corroborate our quantitative findings but, more importantly, indicate a strong human preference for Cut2Next's outputs in terms of overall cinematic continuity and faithful execution of specified editing.

\section{Related Work}
\subsection{Multi-Shot Generation}
Multi-shot generation aims to create sequences of related visual outputs, either as video or sets of images, emphasizing inter-shot consistency for narrative coherence or visual continuity.
In multi-shot video generation, efforts focus on synthesizing continuous events~\cite{lct, mask2dit, shotadapter, vgot,videostudio,atzmon2024multi,bansal2024talc,chen2025skyreels}. Some methods directly use keyframes~\cite{storydiffusion, vgot, one_prompt_one_story,tewel2024training} for image-to-video (I2V) synthesis~\cite{dynamicrafter,i2vgenxl,bar2024lumiere,blattmann2023stable,kong2024hunyuanvideo,yang2024cogvideox}. VideoStudio~\cite{videostudio} leverages entity embeddings for appearance. VGoT~\cite{vgot} uses identity-preserving embeddings for character consistency. MovieDreamer~\cite{moviedreamer} renders visual tokens into I2V keyframes. SynCamMaster~\cite{SynCamMaster} generates 3D-consistent Multi-Angle shots for subsequent editing, while Long Context Tuning~\cite{lct} expands pre-trained models' context to learn scene-level consistency directly.
Alternatively, multi-shot image generation~\cite{storyanchors} often produces "story-frames" for controlled narrative development. IC-LoRA~\cite{iclora} leverages in-context generation with LoRA tuning to achieve consistent environments and characters across frames. CineVerse \cite{cineverse} enhanced IC-LoRA by incorporating detailed annotations for user-controlled shot scale. While methods like StoryDiffusion~\cite{storydiffusion} and One Prompt One Story~\cite{one_prompt_one_story} offer flexibility, they result in character-centric, repetitive perspectives that may weaken narrative impact.
Despite advancements in character and environment consistency across these approaches, they often do not explicitly model the complex editing patterns and cinematic language crucial to professional narrative filmmaking.

\subsection{Subject-driven Generation}
Subject-driven generation aims to create new content featuring specific user-provided subjects.
In image generation, methods like Dreambooth~\cite{dreambooth}, Textual Inversion~\cite{textual_inversion}, and LoRA~\cite{lora} enable subject customization via parameter tuning. Others, like IP-Adapter~\cite{ip_adapter}, utilized external image encoders to inject subject appearance without per-subject fine-tuning. For DiT architectures, methods~\cite{ace++,iclora,ominicontrol,li2025visualcloze,wu2025less} such as IC-LoRA and Ominicontrol,  demonstrated the transformer's inherent capacity for image referencing. 
Extending this to video customization, early works~\cite{id_animator,videobooth,dreamvideo} largely focused on single-concept scenarios. While progress has been made, such as with ConceptMaster~\cite{conceptmaster} in multi-concept video customization, challenges in robustly handling multiple concepts persist.
Despite these advances in generating specified subjects, current methods often handle simple inputs and compositions, thus addressing limited consistency dimensions. Our work targets comprehensive cinematic continuity, encompassing consistent subjects, environments, color, lighting, and style across shot sequences.

\section{Methodology}

\subsection{Dataset Construction and Annotation}
\label{sec:dataset}

\begin{figure}[htbp]
\centering
\includegraphics[width=0.8\linewidth]{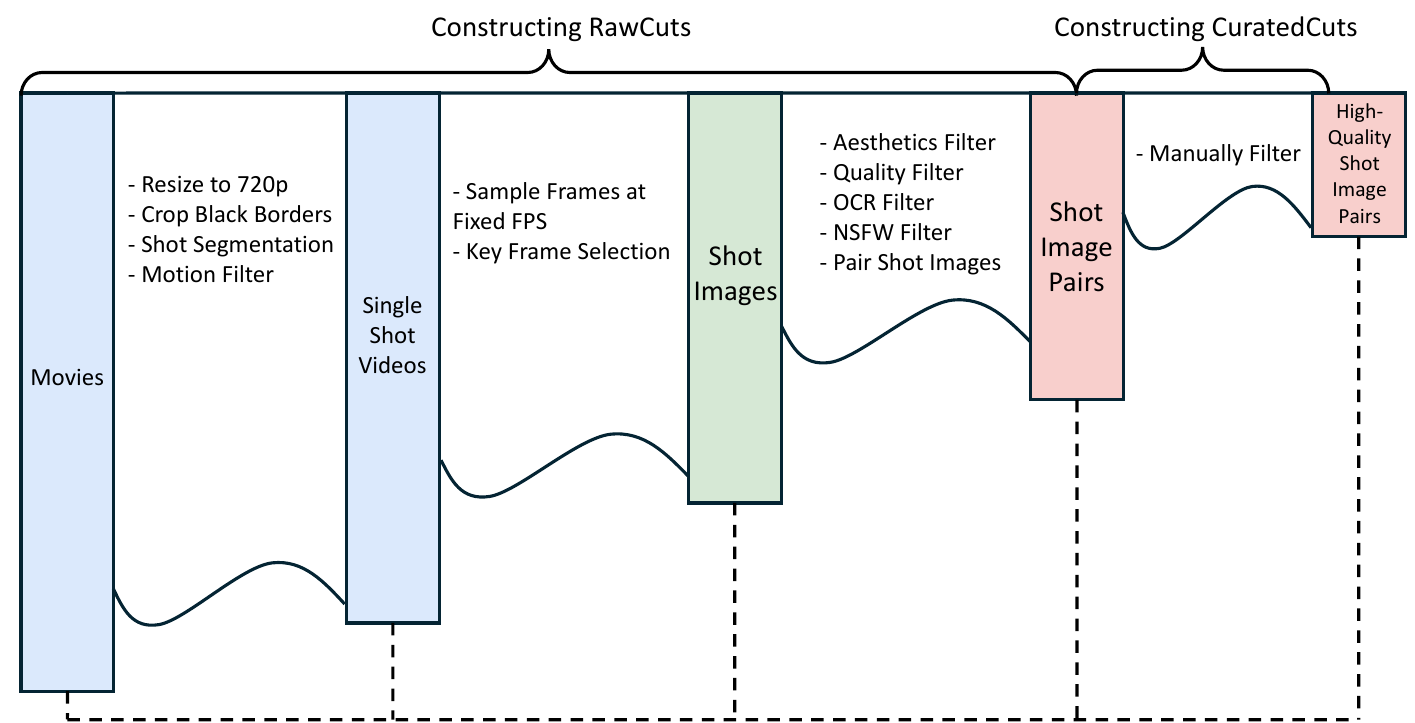}
  \caption{The data construction pipeline for \textbf{RawCuts} and \textbf{CuratedCuts}.}
  \label{fig:data_pipe}
\end{figure}

To generate continuous and cinematically coherent shots, our data strategy employs a two-stage pipeline (Figure~\ref{fig:data_pipe}). First, \textbf{RawCuts}, a large-scale dataset, provides broad exposure to diverse visual transitions for foundational model learning. Second, \textbf{CuratedCuts}, a meticulously selected subset, enables fine-grained refinement of cinematic continuity. Rich textual annotations, via our Hierarchical Prompt Annotation scheme, are then applied to both datasets to capture inter-shot context and per-shot details. This comprehensive annotation strategy enables the model to learn robust visual-semantic correlations, ultimately guiding robust generation even from concise user inputs.

\subsubsection{Dataset Creation Pipeline}
\label{sec:dataset_pipeline}

Our data creation process begins with an automated stage to build the large-scale \textbf{RawCuts} dataset, designed to provide Cut2Next with broad exposure to diverse visual content and transitional patterns. Initially, MovieNet~\cite{movienet} videos are preprocessed by resizing to 720p and removing black borders (\texttt{ffmpeg}~\cite{ffmpeg}). Shot segmentation is performed using TransNetV2~\cite{soucek2020transnetv2}, followed by uniform frame sampling per shot. To ensure quality, we compute aesthetic scores~\cite{laion2022aesthetic} and motion scores (VMAF~\cite{ffmpeg}). We retain only shots with low motion scores to ensure satisfactory per-frame image quality and composition. The frame with the highest aesthetic score from each shot is selected as a keyframe candidate. These candidates undergo further filtering based on aesthetics, image quality (MUSIQ~\cite{musiq}), absence of overlaid text (OCR~\cite{easyocr}), and NSFW content~\cite{nsfw}. Finally, RawCuts is formed by sequentially pairing adjacent keyframes, yielding over 200k shot pairs.

Subsequently, to facilitate the learning of cinematically sophisticated transitions, we manually curate a smaller, high-quality subset from RawCuts, termed \textbf{CuratedCuts}. This meticulous selection prioritizes thousands of shot pairs exhibiting strong adherence to cinematic continuity principles and diverse professional editing techniques (e.g., cut-ins, shot/reverse shots). Emphasis is placed on meaningful character interactions, expressive emotional cues, and varied cinematic styles, making CuratedCuts ideal for refining Cut2Next's ability to generate cinematically coherent shots.

\begin{figure}[htbp]
\centering
\includegraphics[width=0.8\linewidth]{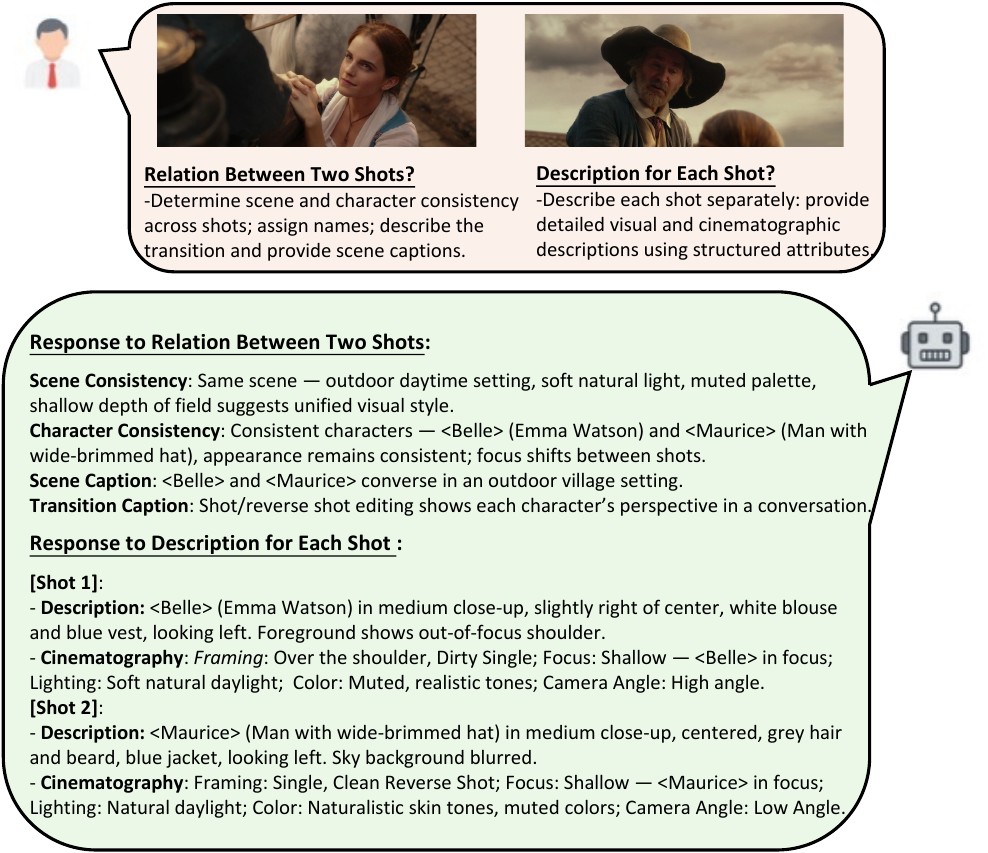}
  \caption{Example of annotating one shot image pair by our Hierarchical Prompt Annotation.}
  \label{fig:data_anotation}
\end{figure}

\subsubsection{Hierarchical Prompt Annotation}
\label{sec:prompt_annotation}

To implement our Hierarchical Multi-Prompting strategy, we adopt an automated annotation process for each image pair ($S_{cond}, S_{tgt}$) in our datasets using the Gemini-2.0-flash~\cite{team2023gemini}. This process generates two complementary types of prompts designed to provide comprehensive textual guidance: \textit{relational prompts ($P^{rel}$)} and \textit{individual prompts ($P^{ind}$)}. The two-tiered prompting scheme captures both the overarching inter-shot narrative relationships and the fine-grained per-shot cinematic details.

\noindent\textbf{Relational Prompts ($P^{rel}$).} $P^{rel}$ encapsulates the semantic and cinematic linkage between $S_{cond}$ and $S_{tgt}$. It includes a high-level overview of their shared visual context (scene, key characters in transition), a narrative interpretation of the shot transition (including editing techniques like shot/reverse shot or cutaways), and evaluations of scene-level (spatial, temporal, stylistic) and character (appearance, attire) continuity. An example is shown in Figure~\ref{fig:data_anotation}.

\noindent\textbf{Individual Prompts ($P^{ind}$).} For each shot ($S_{cond}$ and $S_{tgt}$) independently, $P^{ind}$ offers a detailed description of its visual content and cinematographic characteristics~\cite{shotbench}. This comprises: (1) a concise summary (subject, setting), a detailed description of visual content (appearance, posture, expression, costume, background); and (2) a structured set of cinematography attributes (e.g., shot size/framing, composition, camera angle, focal length). This schema provides a fine-grained representation of visual language. To enhance robustness against incomplete real-world descriptions, the detailed description and attributes are subject to a 20\% dropout rate during training. An example is provided in Figure~\ref{fig:data_anotation}.

This two-stage dataset construction, coupled with our hierarchical annotation strategy, provides a robust foundation for training Cut2Next. Initial training on RawCuts fosters broad visual understanding, while subsequent refinement on CuratedCuts, guided by these detailed prompts, specializes the model for high-quality, cinematically sophisticated next shot generation.

\begin{figure*}[t]
\includegraphics[width=\linewidth]{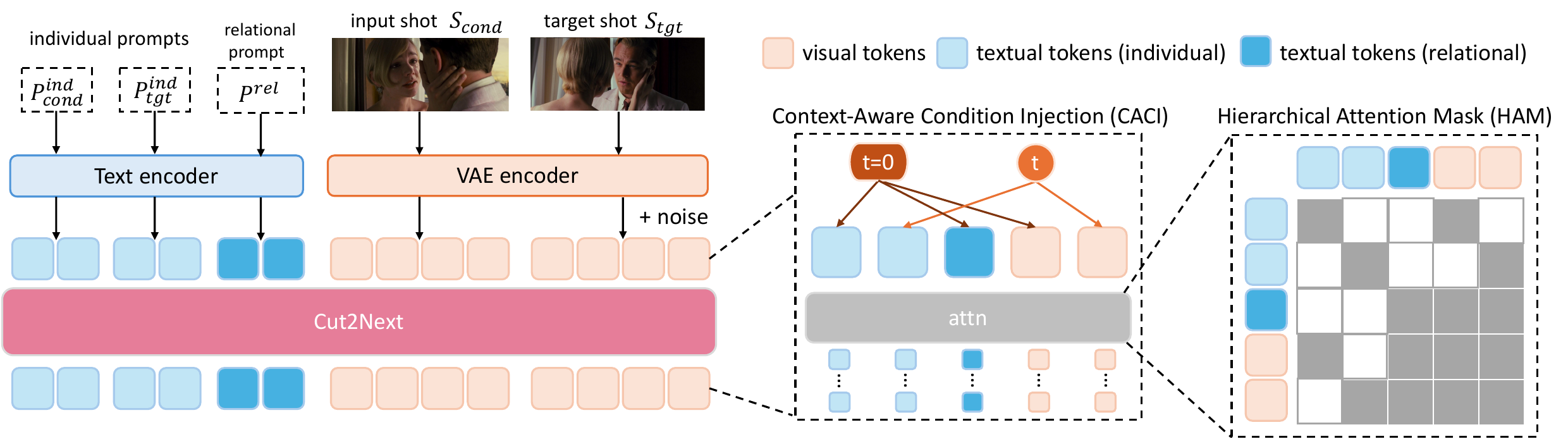}
  \caption{Architecture of Cut2Next. Individual prompts ($P^{ind}_{cond}, P^{ind}_{tgt}$) and a relational prompt ($P^{rel}$) are converted to textual embeddings by a shared text encoder. The conditional shot $S_{cond}$ is encoded by a VAE into clean latents, while the target shot $S_{tgt}$ is encoded and noised for training. These textual and visual tokens form the input to the Cut2Next (DiT-based) model. Our Context-Aware Condition Injection (CACI) module (center right) applies distinct conditioning to AdaLN layers based on token type. The Hierarchical Attention Mask (HAM) (far right) further refines information flow by defining specific attention patterns between different token segments.}
  \label{fig:arch}
\end{figure*}

\subsection{Cut2Next}
To enable the generation of continuous and coherent multi-shot sequences, we leverage the strong generative capabilities of \texttt{FLUX.1-dev}~\cite{flux}, a state-of-the-art Diffusion Transformer (DiT) model. This foundation underpins our proposed method, Cut2Next, a novel framework for synthesizing subsequent shots. The design and operational principles of Cut2Next are elaborated in this section. The overview of Cut2Next is illustrated in Figure \ref{fig:arch}.

\subsubsection{Leveraging Flux via Parameter-Free Modification}

Traditional Text-to-Image (T2I) models iteratively denoise Gaussian noise to generate images. 
The VAE encoder $\mathcal{E}(\cdot)$ first maps the target image $S_{tgt}$ to its latent $z^0_{tgt}$. 
Gaussian noise for timestep $t$ is then added, yielding the noisy latent $z^{t}_{tgt}$. 
This $z^{t}_{tgt}$ is concatenated with the text conditioning $c$ to form the DiT model's input.

To incorporate the conditional shot $S_{cond}$, we also use the pre-trained VAE encoder $\mathcal{E}(\cdot)$ to map
$S_{cond}$ to its latent $z_{cond}=\mathcal{E}(S_{cond})$. 
This ensures $z_{cond}$ is in the same latent space as $z^0_{tgt}$, maintaining compatibility without new encoders. 
$z_{cond}$ is then integrated with the noisy target latent $z^t_{tgt}$ and the primary text conditioning $c$.

Furthermore, Cut2Next uses a Hierarchical Multi-Prompting strategy for textual guidance. 
All prompts are processed by a shared, pre-trained text encoder $\mathcal{T}(\cdot)$ (e.g., T5 \cite{t5}): (1) A relational prompt $P^{rel}$ is encoded to $c^{rel} = \mathcal{T}(P^{rel})$; (2) Individual prompts $P^{ind}_{cond}$ and $P^{ind}_{tgt}$ result in $c^{ind}_{cond} = \mathcal{T}(P^{ind}_{cond})$ and $c^{ind}_{tgt} = \mathcal{T}(P^{ind}_{tgt})$.
These textual embeddings and visual latents are concatenated to form the DiT model's input sequence $z_{model}$:
\begin{equation}
\label{eqn:dit_input_final}
z_{model} = \text{concat}(c^{rel}, c^{ind}_{cond}, c^{ind}_{tgt}, z_{cond}, z^{t}_{tgt}).
\end{equation}
This rich input $z_{model}$ allows the DiT to condition on visual context from $S_{cond}$ and multi-level textual instructions, efficiently reusing existing encoders.

To adapt shared DiT blocks for this complex, multi-modal input, lightweight LoRA fine-tuning \cite{lora} is applied. 
This enables the model to learn the interplay between these conditioning signals without full-parameter updates.

\subsubsection{Context-Aware Condition Injection (CACI)}

Standard Flux uses AdaLN-Zero \cite{dit} with synchronously noised visual tokens and a unified textual context ($c^{pool}$), all conditioned by timestep $t$. 
Our Hierarchical Multi-Prompting, however, introduces heterogeneous inputs: noise-free conditional visual latents ($z_{cond}$), noisy target visual latents ($z^{t}_{tgt}$), and multiple distinct textual embeddings ($c^{rel}, c^{ind}_{cond}, c^{ind}_{tgt}$). 
This departure from standard conditioning demands a nuanced approach.

We propose \textbf{Context-Aware Condition Injection (CACI)}. 
CACI enables DiT blocks to be \textit{aware} of each token segment's context and role. 
It then \textit{tailors} AdaLN-Zero inputs for each segment. Specifically, CACI differentiates visual latent conditioning: (1) Noise-free $z_{cond}$ tokens are modulated by AdaLN layers. These use $t=0$ (reflecting their clean state) and context $c^{ind\_pool}_{cond}$ (from $P^{ind}_{cond}$). (2) Noisy $z^{t}_{tgt}$ tokens use the diffusion timestep $t$ and context $c^{ind\_pool}_{tgt}$ (from $P^{ind}_{tgt}$).
This allows specialized processing for $z_{cond}$ and $z^{t}_{tgt}$.

CACI also applies context-awareness to textual tokens. 
Their AdaLN conditioning aligns with their semantic scope. 
Individual textual tokens $c^{ind}_{cond}$ and $c^{ind}_{tgt}$ mirror the timesteps of their visual counterparts ($t=0$ for $z_{cond}$; diffusion $t$ for $z^{t}_{tgt}$). 
For relational tokens $c^{rel}$, empirical results (Figure~\ref{fig:loss}) showed that $t=0$ yielded a lower initial loss than diffusion $t$, despite similar final convergence. 
Thus, we use $t=0$ for $c^{rel}$, treating it as part of the "initial, clean context" with $z_{cond}$. In essence, CACI's context-aware, token-type-specific conditioning within AdaLN-Zero layers effectively manages our strategy's heterogeneous conditioning.

\subsubsection{Hierarchical Attention Mask (HAM) for Structured Interactions}
Our composite input $z_{model}$ requires structured information flow in DiT blocks. 
Standard full self-attention might undesirably mix distinct conditioning signals, diluting their influence. 
For example, individual prompts should guide their corresponding visual segments. The relational prompt should mediate between visual components. Direct cross-interference between disparate textual cues must be avoided.

We introduce the \textbf{Hierarchical Attention Mask (HAM)}, a predefined, non-learnable binary mask for self-attention. 
HAM selectively controls attention between token types. 
This structures information exchange according to our Hierarchical Multi-Prompting and CACI, upholding textual prompt independence. HAM enforces specific attention pathways:

\textbf{- Visual Dynamics:} Conditional ($z_{cond}$) and target ($z^{t}_{tgt}$) visual tokens mutually attend, besides intra-segment self-attention.

\textbf{- Localized Individual Prompts:}

\textbf{(1)} Conditional text ($c^{ind}_{cond}$) attends only to itself and $z_{cond}$; $z_{cond}$ also attends back to $c^{ind}_{cond}$.

\textbf{(2)} A similar isolated scope exists for target text ($c^{ind}_{tgt}$) and $z^{t}_{tgt}$.

\textbf{(3)} Attention is masked between an individual text prompt and non-corresponding visual segments or other text prompts (including $c^{rel}$). This ensures $P^{ind}_{cond}$ and $P^{ind}_{tgt}$ exclusively modulate their designated visual counterparts without cross-contamination.

\textbf{- Relational Prompt Bridging:}
        
\textbf{(1)} Relational text ($c^{rel}$) performs self-attention. Its cross-modal interaction is with \textit{both} $z_{cond}$ and $z^{t}_{tgt}$ (and vice-versa). This broad visual scope is crucial for $c^{rel}$ to establish inter-shot relationships.

\textbf{(2)} $c^{rel}$ is masked from attending to $c^{ind}_{cond}$ or $c^{ind}_{tgt}$, and vice-versa, maintaining textual independence.

By enforcing these pathways, HAM enables Cut2Next to disentangle and leverage multi-level guidance. 
Each signal contributes precisely to synthesizing $S_{tgt}$ relative to $S_{cond}$, without undesired textual cross-talk.

\section{Experiments}
\subsection{Experimental Setup}

\noindent\textbf{Implementation details.}
We adopt \texttt{FLUX.1-dev} as our base model, with LoRA \cite{lora} layers set to a rank of 256. 
The Adam \cite{adam} optimizer is employed with a learning rate of $1 \times 10^{-4}$. 
Our model is trained on shot images with varying aspect ratios, at a resolution of approximately 1K pixels (e.g., $1024 \times 576$).
Our two-stage training strategy begins with pre-training the model on the RawCuts Dataset for 2 epochs. This uses an accumulated batch size of 64 on 8$\times$A100 GPUs. 
Subsequently, the model is fine-tuned for 2500 iterations on the CuratedCuts Dataset. 

\noindent\textbf{Baselines.}
Cut2Next is pioneering work in the domain of conditional shot generation. To date, no publicly available open-source baselines specifically designed for generating a subsequent shot conditioned on a preceding one exist for direct comparison. Therefore, we adapted the IC-LoRA \cite{iclora} framework, originally designed for text-to-multiple-image generation, to serve as a strong baseline for our task. We term this adapted version \textbf{IC-LoRA-Cond}.

The original IC-LoRA generates multiple shots from a textual description using markers like \texttt{[IMAGE1]}, \texttt{[IMAGE2]}. 
To adapt it for conditional shot generation, we modify the first shot's input: we directly use the clean latent of the provided conditional shot image $S_{cond}$. 
The subsequent target shot is then generated conditioned on the $S_{cond}$ latents. 
Loss is calculated only on the target shot's noisy latents. 
Textual input uses IC-LoRA's concatenated prompt: an overall description plus separate descriptions for the conditional (\texttt{[IMAGE1]}) and target (\texttt{[IMAGE2]}) shots.

\noindent\textbf{Evaluation.}
We evaluate Next Shot Generation using standard automatic metrics for visual consistency and textual fidelity.
Regarding visual consistency, we measure the cosine similarity between the input shot and generated shot embeddings. This uses two feature spaces: CLIP \cite{clip} feature space (CLIP-I Similarity) and DINO \cite{dino} feature space (DINO Similarity). 
Regarding textual fidelity, we assess alignment of the generated shot with its individual prompt ($P^{ind}_{tgt}$), we compute the cosine similarity between its CLIP image embedding and the prompt's CLIP text embedding (CLIP-T Fidelity).
To measure perceptual similarity to real cinematic choices, we evaluate against the ground-truth (GT) shots using Fréchet Inception Distance (FID).
All evaluations use our new CutBench benchmark. CutBench contains hundreds of diverse movie shot images with our hierarchical prompts designed to elicit specific cinematic continuations that test both continuity and varied editing. 

\begin{figure*}[htbp]
\centering
\includegraphics[width=\linewidth]{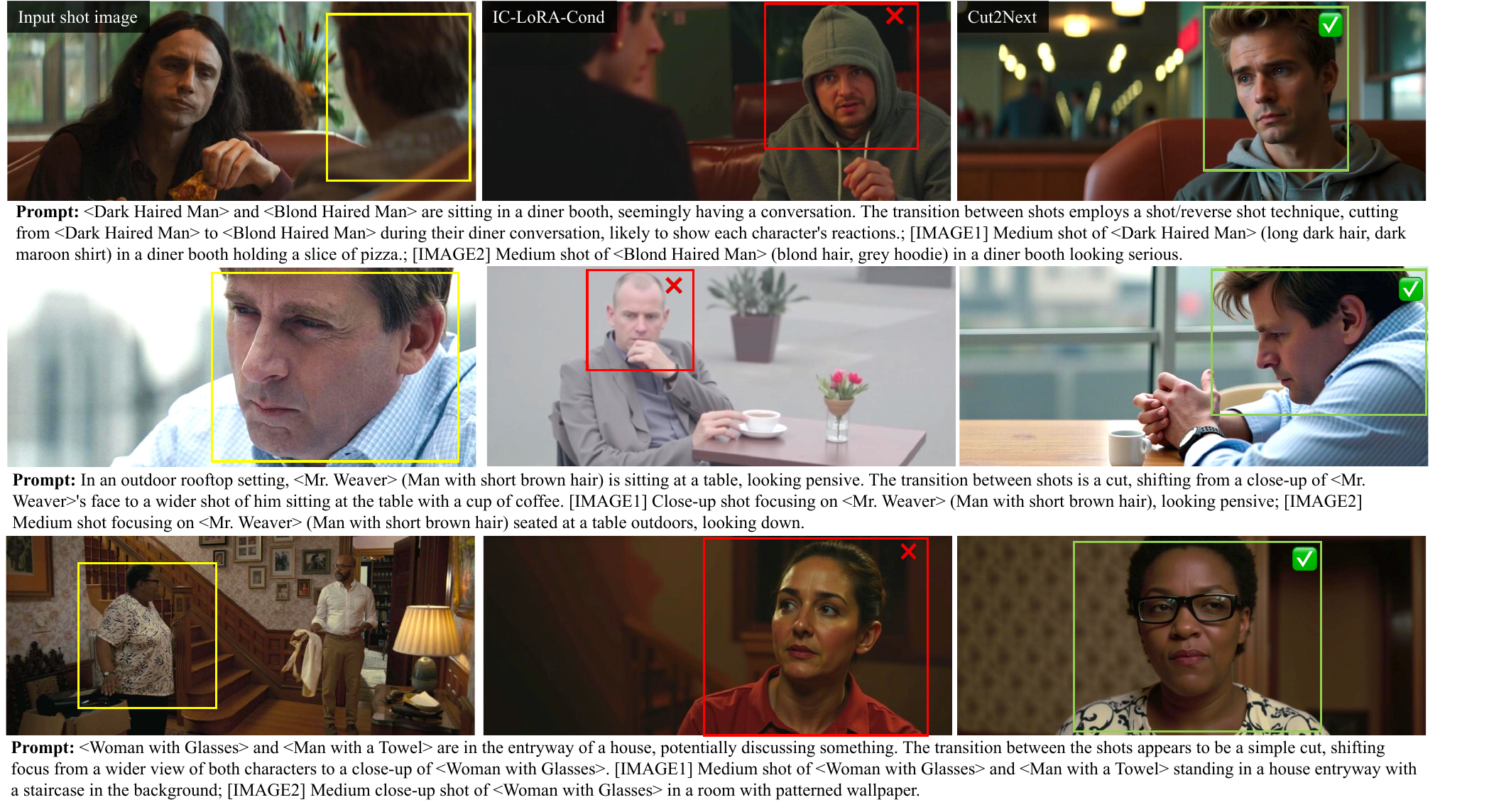}
  \caption{Visual comparison of Cut2Next and IC-LoRA-Cond \cite{iclora}.}
  \label{fig:vs_iclora}
\end{figure*}

\subsection{Main results}

\noindent\textbf{Qualitative analyses.}
We compare Cut2Next with IC-LoRA-Cond \cite{iclora}. Figure~\ref{fig:vs_iclora} empirically demonstrates Cut2Next's significant advantages in next shot generation. For instance, in a diner conversation (Row 1), Cut2Next generates a shot/reverse shot maintaining character identity ("Blond Haired Man"), crucial for dialogue. IC-LoRA-Cond fails, introducing a dissimilar individual and disrupting narrative flow. 
Similarly, for "Mr. Weaver" (Row 2), Cut2Next logically continues focus with appropriate framing. IC-LoRA-Cond again introduces an unrelated figure. 
In an indoor setting (Row 3), Cut2Next successfully transitions from a wider view to a close-up of the "Woman with Glasses," maintaining focus and continuity. IC-LoRA-Cond's output deviates significantly.
This visual comparison affirms Cut2Next's enhanced capacity for generating temporally coherent, character-consistent, and cinematically plausible subsequent shots.

\noindent\textbf{Quantitative evaluations.}
As shown in Table~\ref{tab:vs_iclora}, our quantitative evaluation on CutBench demonstrates Cut2Next's superior performance over IC-LoRA-Cond across all metrics. For inter-shot visual consistency, Cut2Next achieves higher DINO Similarity (0.4952 vs. 0.4669) and CLIP-I Similarity (0.7298 vs. 0.7152), indicating greater visual continuity with the input shot. For textual fidelity, Cut2Next also shows enhanced alignment, with a CLIP-T Fidelity of 0.2979 compared to IC-LoRA-Cond's 0.2805. Besides, our method achieves a significantly lower FID (59.37) than the baseline (80.43). These metrics collectively affirm that Cut2Next generates shots that are more visually consistent with the preceding context and more faithful to textual descriptions.

\begin{table}[htbp]
  \centering
  \small
   \caption{Performance comparison between Cut2Next and IC-LoRA-Cond on CutBench. Best scores are in \textbf{bold}.}
  \label{tab:vs_iclora}
  \setlength{\tabcolsep}{8pt}
  \resizebox{0.5\textwidth}{!}{%
    \begin{tabular}{lcccc}
      \toprule
       & DINO~$\uparrow$    & CLIP-I~$\uparrow$  & CLIP-T~$\uparrow$ & FID~$\downarrow$ \\ 
      \midrule
      IC‐LoRA-Cond
        & $0.4669$ 
        & $0.7152$ 
        & $0.2805$
        & $80.43$ \\  
      \textbf{Cut2Next (ours)}
        & $\mathbf{0.4952}$ 
        & $\mathbf{0.7298}$ 
        & $\mathbf{0.2979}$ 
        & $\mathbf{59.37}$ \\
      \bottomrule
    \end{tabular}%
  }
\end{table}

\subsection{Ablation study}

\noindent\textbf{Effectiveness of two-stage training.}
To validate our two-stage strategy, we compared our full "Cut2Next (ours)" with variants fine-tuned solely on RawCuts or CuratedCuts (Table~\ref{tab:vs_different_dataset}).
The RawCuts-only model achieved DINO 0.4853, CLIP-I 0.7135, and CLIP-T 0.2951. The smaller CuratedCuts-only model performed better (DINO 0.4913, CLIP-I 0.7221, CLIP-T 0.2973), highlighting data quality's impact, especially with a strong base like \texttt{FLUX.1-dev}.
Our two-stage "Cut2Next (ours)" yields the best results, validating our approach. This shows a synergistic benefit: RawCuts pre-training builds a broad foundation, refined by CuratedCuts fine-tuning, balancing robustness with specialized performance.
\begin{table}[htbp]
  \centering
  \small
  \caption{Cut2Next ablation on training datasets. DINO, CLIP-I, and CLIP-T scores. Higher is better. Best in \textbf{bold}.}
  \label{tab:vs_different_dataset}
  \setlength{\tabcolsep}{8pt}
  \resizebox{0.5\textwidth}{!}{%
    \begin{tabular}{lccc}
      \toprule
       & DINO~$\uparrow$    & CLIP-I~$\uparrow$  & CLIP-T~$\uparrow$  \\ 
      \midrule
      Cut2Next (RawCuts-only)
        & $0.4853$ 
        & $0.7135$ 
        & $0.2951$ \\
      Cut2Next (CuratedCuts-only)
        & $0.4913$ 
        & $0.7221$ 
        & $0.2973$ \\
      \textbf{Cut2Next (ours)}
        & $\mathbf{0.4952}$ 
        & $\mathbf{0.7298}$ 
        & $\mathbf{0.2979}$ \\
      \bottomrule
    \end{tabular}%
  }
\end{table}

Figure~\ref{fig:wo_pretraining} visually confirms this. Our full model captures challenging attributes ("green highlights," top row) and complex scenarios ("injured man," "SECURITY GUARD," bottom row). The variant fine-tuned only on CuratedCuts (w/o pretraining) fails on these specifics. This demonstrates that while CuratedCuts refines continuity, RawCuts pre-training provides crucial robustness for complex or less common visual elements.



\noindent\textbf{Effectiveness of Hierarchical Multi-Prompting strategy.}
To validate our Hierarchical Multi-Prompting strategy, we ablated the relational prompt ($P^{rel}$), comparing our full Cut2Next model against a variant using only individual prompts ("Cut2Next w/o relational prompt"), with adapted CACI and HAM.
Table~\ref{tab:wo_relational_prompt} shows that removing $P^{rel}$ significantly degrades inter-shot visual consistency: our full model achieved higher DINO (0.4952 vs. 0.4752) and CLIP-I Similarity (0.7298 vs. 0.7140) scores. This demonstrates $P^{rel}$'s crucial role in generating visually coherent subsequent shots. Interestingly, text fidelity (CLIP-T) showed a marginal decrease for the full model (0.2979) compared to the ablated version (0.2984). This suggests $P^{rel}$, while significantly enhancing visual continuity, might introduce a subtle trade-off or different emphasis in overall textual adherence.
\begin{table}[htbp]
  \centering
  \small
   \caption{Comparison of Cut2Next and Cut2Next without relational prompt. DINO, CLIP-I, and CLIP-T scores. Higher is better. Best in \textbf{bold}.}
  \label{tab:wo_relational_prompt}
  \setlength{\tabcolsep}{8pt}
  \resizebox{0.5\textwidth}{!}{%
    \begin{tabular}{lccc}
      \toprule
       & DINO~$\uparrow$    & CLIP-I~$\uparrow$  & CLIP-T~$\uparrow$  \\ 
      \midrule
      Cut2Next (w/o relational prompt)
        & $0.4752$ 
        & $0.7140$ 
        & $\mathbf{0.2984}$ \\
      \textbf{Cut2Next (ours)}
        & $\mathbf{0.4952}$ 
        & $\mathbf{0.7298}$ 
        & $0.2979$ \\
      \bottomrule
    \end{tabular}%
  }
\end{table}

Qualitative results in Figure~\ref{fig:wo_relational_prompt} corroborate these findings. Without $P^{rel}$ (right column), issues like character identity loss (e.g., second row) and failure to follow complex editing instructions like "shot/reverse shot" (first row) become apparent, even with individual prompts. While this ablated version can still match scene environment and tone better than simpler baselines (e.g., IC-LoRA-Cond), it clearly struggles with crucial aspects of cinematic continuity, underscoring the importance of $P^{rel}$.

\begin{figure}[t]
\centering
\includegraphics[width=0.7\linewidth]{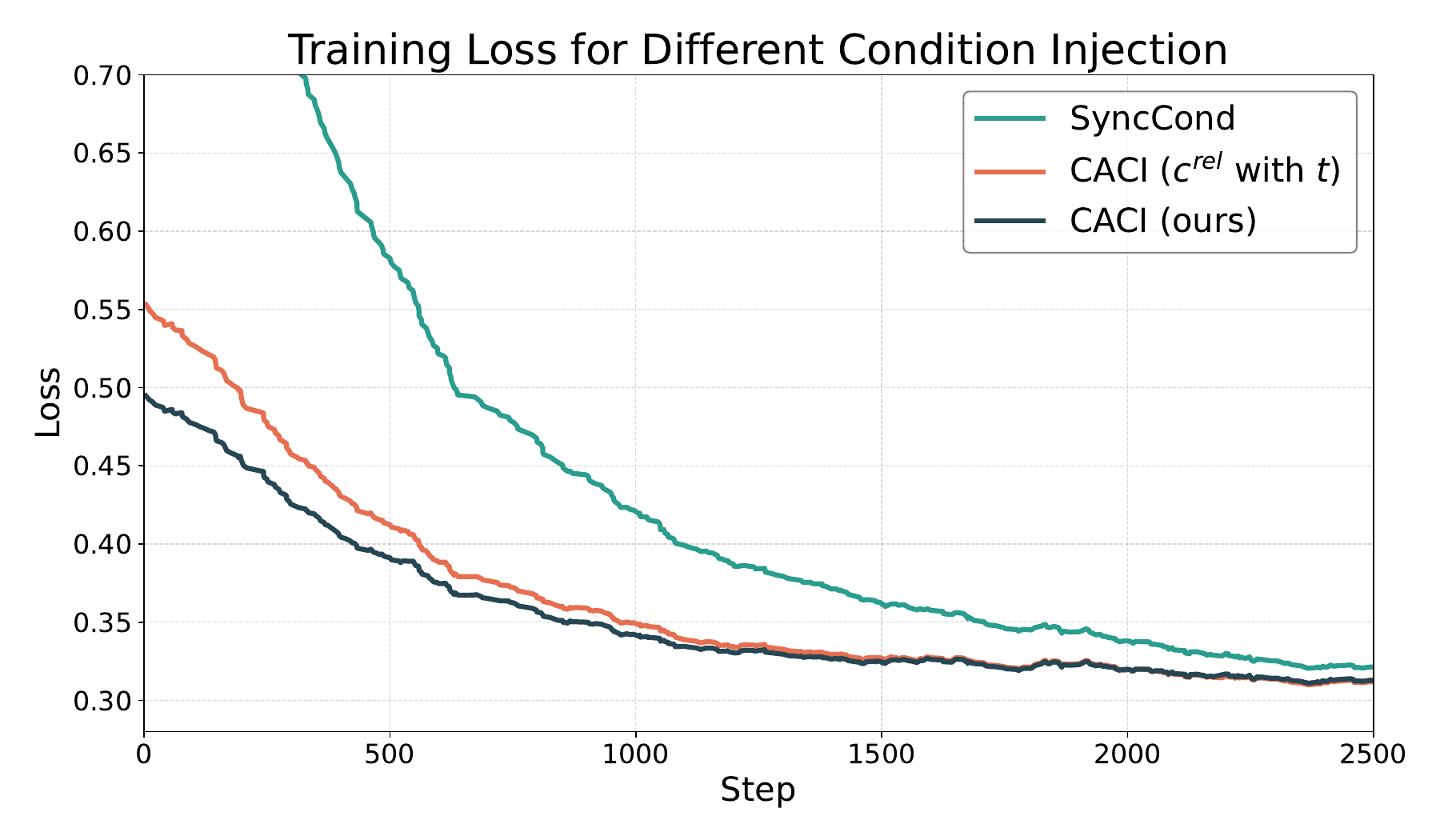}
  \caption{Training loss comparison on RawCuts. CACI (ours) shows improved convergence over SyncCond and CACI ($c^{rel}$ with $t$).}
  \label{fig:loss}
\end{figure}

\noindent\textbf{Effectiveness of Context-Aware Condition Injection.}
To validate our Context-Aware Condition Injection (CACI) design, we ablated its components. 
We compared CACI against:
1.Synchronous Conditioning (SyncCond): A baseline where all visual and textual tokens are conditioned with the current diffusion timestep $t$, similar to the original Flux paradigm; 2. CACI ($c^{rel}$ with $t$): A CACI variant where relational textual tokens ($c^{rel}$) use the current timestep $t$ instead of $t=0$ (as in our main CACI model).
Figure~\ref{fig:loss} shows training loss dynamics on RawCuts. 
SyncCond exhibited the highest initial loss and slowest convergence, suggesting inefficiency when processing noise-free conditions at non-zero timesteps. 
In contrast, both CACI variants showed faster initial convergence and lower overall loss. 
Notably, our proposed CACI ($c^{rel}$ with $t=0$) achieved a slightly lower initial loss than CACI ($c^{rel}$ with $t$), though both converged similarly. 
This supports our $t=0$ choice for $c^{rel}$, indicating better initial optimization and underscoring the benefit of context-specific timestep conditioning for efficient training.

\subsection{Perceptual User Study}
\label{sec:userstudy}

To complement automatic evaluation, we conduct a human-centric study to assess whether Cut2Next produces perceptually better results than the IC-LoRA-Cond baseline. The evaluation focuses on two criteria central to our work: (1) Cinematic Continuity (consistency of characters, environment, and key visual details from the prior shot).
(2) Adherence to Editing (accurate execution of the intended cut/transition type).

We recruited fifteen participants, including nine graduate students specializing in visual or multimedia fields, three professional video editors, and three lay users with no formal training. All participants reported normal or corrected-to-normal vision and voluntarily provided informed consent. We randomly select 100 samples from the CutBench test split and generated paired outputs from Cut2Next and the IC-LoRA-Cond for each prompt. Participants are presented with the complete input context, including both the text prompt and the corresponding condition shot, and asked to select the preferred result according to two criteria: cinematic quality, and adherence to editing. Each image pair is shown side-by-side in a randomized order, with no option for ties. This process is repeated 4 times to compute the standard deviation. For each criterion, we calculate the average preference rate of each method.

As shown in Table~\ref{tab:userstudy}, Cut2Next is overwhelmingly preferred across all criteria, demonstrating significant perceptual advantages over IC-LoRA-Cond.

\begin{table}[htbp]
    \centering
    \small
    \caption{Human preference (\%) between Cut2Next and IC-LoRA-Cond~\cite{iclora}. Higher is better and best is in \textbf{bold}.}
    \label{tab:userstudy}
    \setlength{\tabcolsep}{4pt}
    \begin{tabular}{lccc}
        \toprule
        Method & Cinematic Continuity $\uparrow$ & Adherence to Editing $\uparrow$ \\
        \midrule
        IC-LoRA-Cond~\cite{iclora}    & $6.3 \pm 1.8$& $3.5 \pm 0.8$\\
        \textbf{Cut2Next (ours)}   & $\mathbf{93.7 \pm 1.8}$ & $\mathbf{96.5 \pm 0.8}$  \\
        \bottomrule
    \end{tabular}
     
\end{table}


\section{Conclusion and limitations}
We introduced Next Shot Generation (NSG), a novel task targeting the critical need for both professional editing patterns and strict cinematic continuity in multi-shot video. Our framework, Cut2Next, leverages a Diffusion Transformer with an innovative Hierarchical Multi-Prompting strategy. Architectural enhancements (CACI and HAM) seamlessly integrate these controls.
Evaluations on our new datasets (RawCuts, CuratedCuts) and benchmark (CutBench) show Cut2Next excels quantitatively. Crucially, user studies confirm a strong preference for Cut2Next's adherence to editing patterns and overall cinematic continuity. Our work focuses on canonical, human-centric editing patterns, thus it is limited to human-centric scenarios. Besides, it may fail in producing action sequences as we filtered high-motion shots to ensure keyframe quality. Long-term coherence is still a critical challenge. A naive autoregressive approach fails for our task because cinematic cuts (e.g., shot/reverse shot) create large visual shifts, leading to character identity loss.

\bibliographystyle{plain}
\bibliography{main}

\appendix

\begin{figure*}[htbp]
\centering
\includegraphics[width=1.\linewidth]{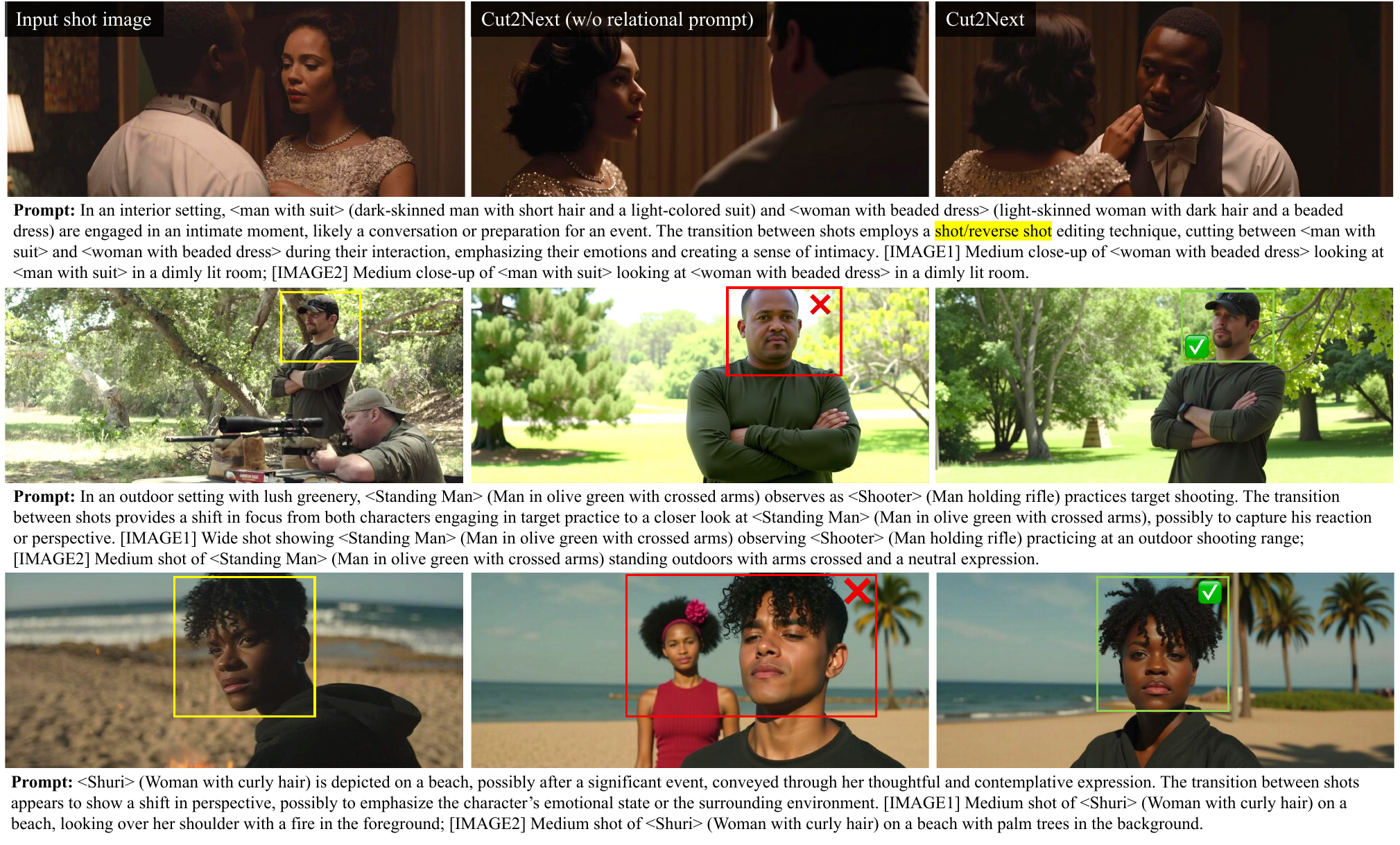}
  \caption{Visual comparison of Cut2Next and Cut2Next (w/o relational prompt).}
  \label{fig:wo_relational_prompt}
\end{figure*}

\begin{figure*}[htbp]
\centering
\includegraphics[width=1.\linewidth]{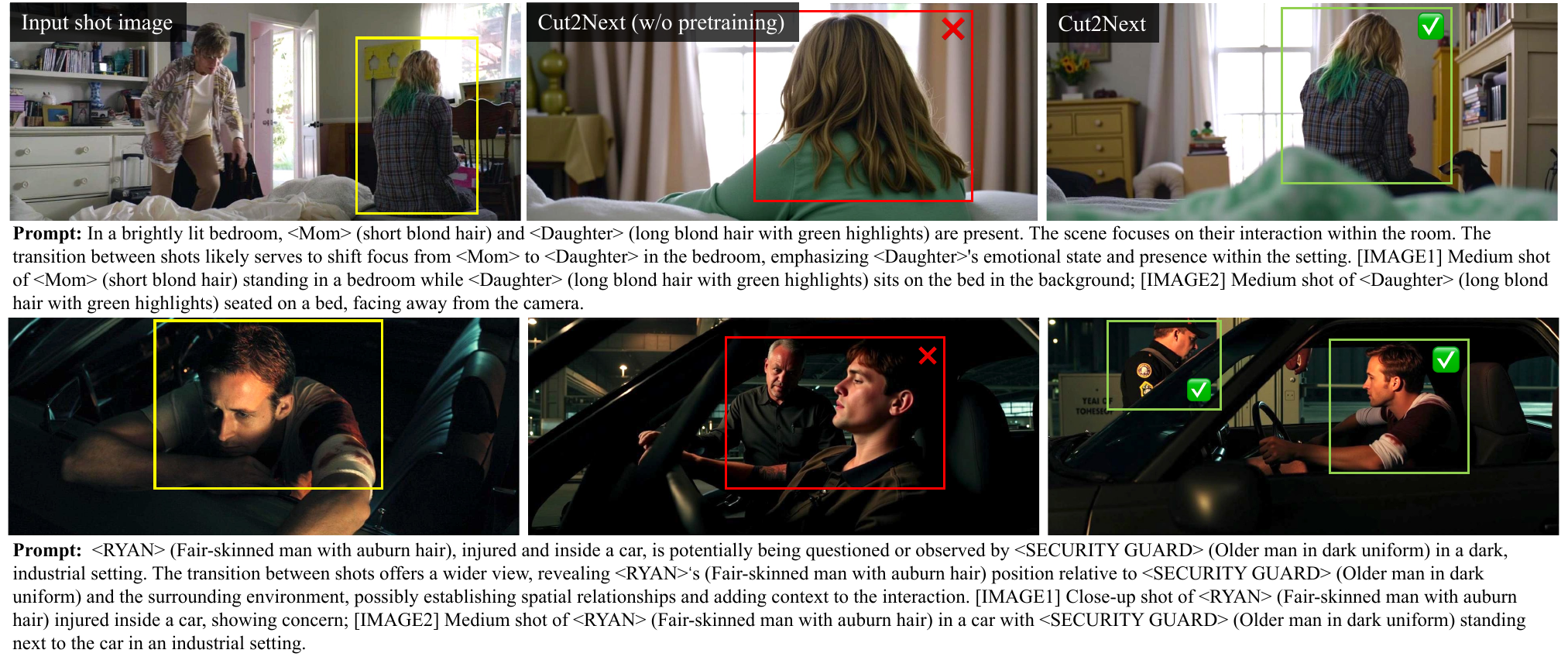}
  \caption{Visual comparison of Cut2Next and Cut2Next (without pretraining on RawCuts dataset).}
  \label{fig:wo_pretraining}
\end{figure*}

\begin{figure*}[htbp]
\centering
\includegraphics[width=1.\linewidth]{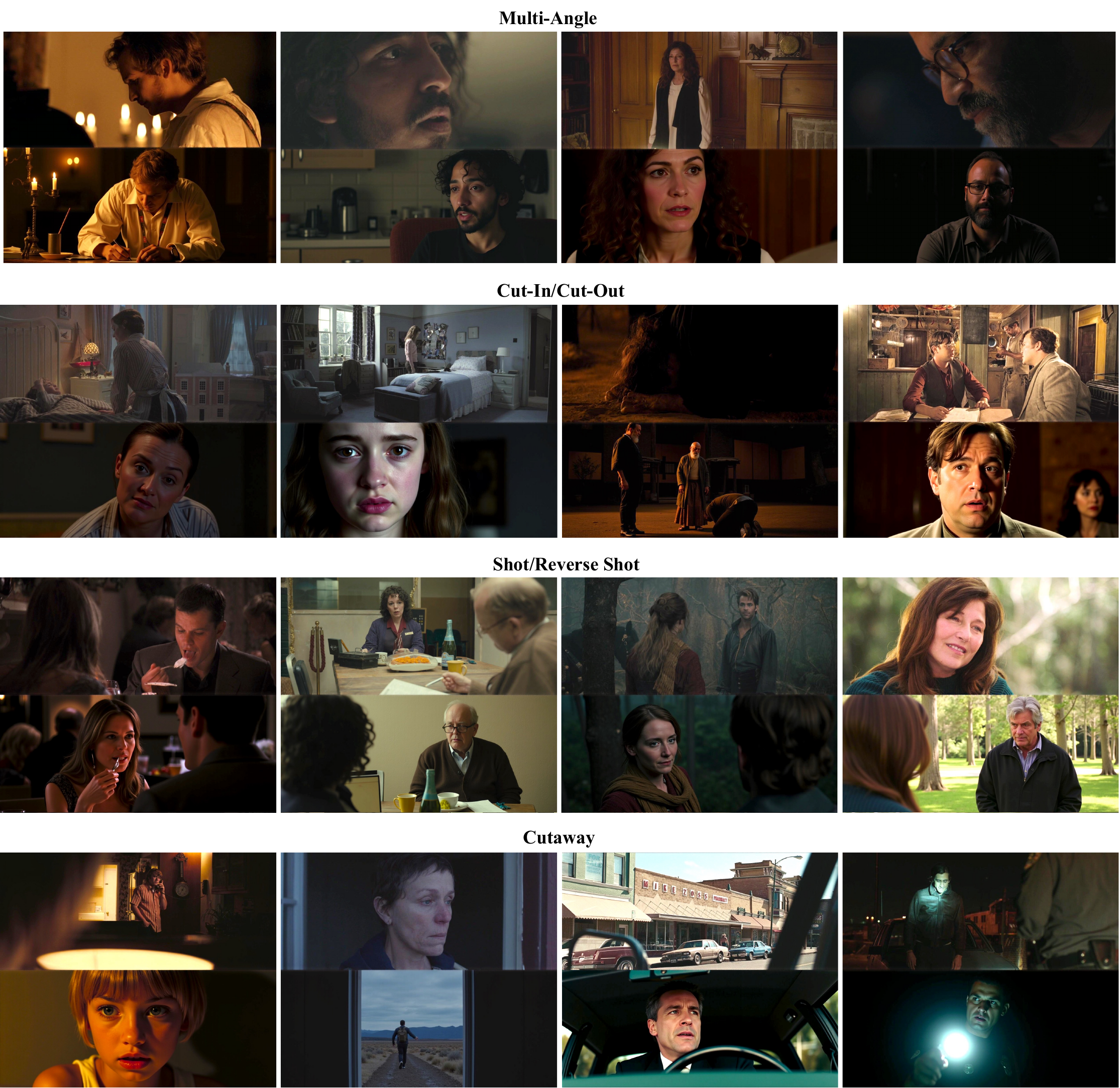}
  \caption{\textbf{More visual results from Cut2Next} demonstrating performance across different editing patterns: Multi-Angle, cut-in/cut-out, shot/reverse shot, and cutaway. For each pair, the input shot image is shown above the generated subsequent shot. Prompts are omitted due to space limitations.}
  \label{fig:more_results}
\end{figure*}

\end{document}